# Distinguishing Parkinson's Patients Using Voice-Based Feature Extraction and Classification

**Research Assistant. Burak ÇELİK**
Kocaeli University / Faculty of Engineering / Electronics and Communication Engineering
*burak.celik@kocaeli.edu.tr, ORCID:0000-0002-3204-5444*

**Associate Professor Dr. Ayhan AKBAL**
Fırat University / Faculty of Engineering / Electrical and Electronics Engineering
*ayhan_akbal@firat.edu.tr, ORCID:* 0000-0001-5385-9781

**Abstract**

Parkinson's disease (PD) is a progressive neurodegenerative disorder that impacts motor functions and speech characteristics This study focuses on differentiating individuals with Parkinson's disease from healthy controls through the extraction and classification of speech features. Patients were further divided into 2 groups. Med On represents the patient with medication, while Med Off represents the patient without medication. The dataset consisted of patients and healthy individuals who read a predefined text using the H1N Zoom microphone in a suitable recording environment at Fırat University Neurology Department. Speech recordings from PD patients and healthy controls were analyzed, and 19 key features were extracted, including jitter, luminance, zero-crossing rate (ZCR), root mean square (RMS) energy, entropy, skewness, and kurtosis**.**These features were visualized in graphs and statistically evaluated to identify distinctive patterns in PD patients. Using MATLAB's Classification Learner toolbox, several machine learning classification algorithm models were applied to classify groups and significant accuracy rates were achieved. The accuracy of our 3-layer artificial neural network architecture was also compared with classical machine learning algorithms. This study highlights the potential of noninvasive voice analysis combined with machine learning for early detection and monitoring of PD patients. Future research can improve diagnostic accuracy by optimizing feature selection and exploring advanced classification techniques.

## 1. INTRODUCTİON

Parkinson's disease (PD), the second most prevalent neurodegenerative disorder globally, impacts over 10 million individuals and arises from the degeneration of dopaminergic neurons in the substantia nigra. Patients with Parkinson's disease (PD) who experience motor fluctuations and dyskinesias often find it challenging to manage their condition with oral therapies (Hemmerling and Wojcik-Pedziwiatr, 2022). Parkinson's disease (PD) is a progressive disorder of the central nervous system that primarily impacts the motor system. (Agarwal, Chandrayan and Sahu, 2016). There are numerous symptoms in the population suffering from the disease, such as tremors, slowness of movements, disturbance of posture and balance, and stiffness of the muscles, but dysphonia (changes in speech and articulation) is the most important antecedent(Vadovský and Paralič, 2017). In(Rana et al., 2022), the use of machine learning algorithms through audio analysis for early detection of Parkinson's disease is described. In this study, various machine learning algorithms that analyze audio features such as Jitter and Shimmer analysis, Average Energy over Time analysis, Average Frequency, FFT-Frequency analysis, Kurtosis, Skewness, Zero Crossing Ratio (ZCR), Root Mean Square (RMS) Entropy, Power Bandwidth are utilized to diagnose Parkinson's disease. Furthermore, the significance of early diagnosis and treatment of Parkinson's disease, as well as the limitations and challenges of existing diagnostic methods, are highlighted. In the study, machine learning algorithms including Naive Bayes, MLP, k-NN, SVM, Logistic Regression, and Random Forest were utilized. The study(Frid et al., 2014) in proposed using machine learning to diagnose Parkinson's disease by analyzing natural speech patterns. In study, data were generated by having 43 patients and 9 healthy individuals read the text titled "Rainbow Passage". In(Sztaho et al., 2017), a method was proposed to predict the severity of Parkinson's disease based on speech rhythm related features extracted from fluent speech. Various machine learning methods were used including Support Vector Machines, Decision Trees and Frandom Forest. Only clip-on microphone was used to create the dataset in the study. The study in(Tougui, Jilbab and Mhamdi, 2020) presented a methodology to distinguish Parkinson's patients from healthy controls using a large sample of 18,210 smartphone recordings. The audio recordings were used to create a dataset containing 80,594 samples and 138 features extracted from time, frequency, and cepstral domains. The dataset was analyzed using basic machine learning models, employing four classifiers: Linear Support Vector Machine (SVM), K-nearest neighbor (k-NN). The study in( Vasquez-Correa et al., 2016) proposed two new features based on word accuracy and dynamic time warping algorithm to evaluate the intelligibility deficits of patients with this disease using an automated speech recognition system by offering different features in this regard. The audio data consists of 50 patients with Parkinson's disease and 50 healthy control patients. The text read to the patients consists of 6 sentences and 36 words. The audio data was collected in a soundproof cabin using a microphone and a professional sound card with a sampling frequency of 44.1 kHz and a resolution of 16 bits. The study(Zhang et al., 2016) proposed an algorithm aimed at enhancing classification accuracy and stability for Parkinson's disease diagnosis using speech samples. The audio data was obtained from a dataset repository website. The dataset is divided into two sets: Training Data and Test Data. The Training Data consists of 40 subjects, including 20 healthy individuals (10 females and 10 males) and 20 patients (6 females and 14 males). The study in (Orozco-Arroyave et al., 2014) presented a novel Spanish speech database designed to analyze individuals with Parkinson's disease. The audio data was recorded in a soundproof booth using a dynamic omnidirectional microphone The database contains recordings of both healthy controls and patients diagnosed with Parkinson's disease. The speech of 50 Parkinson's patients and 50 healthy controls, 25 male and 25 female, is available. SVM also achieved high accuracy in this study. The study in(Little et al., 2009) investigated the suitability of dysphonia measurements for telemonitoring Parkinson's disease. The study aimed to evaluate various measurements, including a new measurement called pitch period entropy (PPE) to distinguish between healthy and Parkinson's disease individuals. Voice recordings were obtained using a head-mounted microphone placed 8 cm from the lips. The recordings were made in a soundproof booth. The dataset consists of 195 continuous phonations from 31 male and female subjects. 23 of these subjects were diagnosed with Parkinson's disease, while the remaining subjects served as healthy controls. This study, like other studies, reported the use of the SVM classifier. In another study(Ulaş, 2023), Another study presented a machine learning approach for detecting Parkinson's disease using various potential acoustic signal features from both Parkinson's patients and control subjects. The audio data used was collected from the Istanbul acoustic database, which includes 252 individuals: 64 controls and 188 Parkinson's patients. The dataset comprised voice information from 122 female subjects (41 controls and 81 Parkinson's patients) and 130 male subjects (41 controls and 81 Parkinson's patients). In the study, fundamental frequency, jitter, shimmer and harmonic acoustic features were extracted from the audio data. These features were selected according to their classification contribution using 3 important feature selection techniques. Modeling was performed using a machine approach that included feature extraction, feature selection and classification.

## 2. MATERİAL AND METHOD

In this study, speech recordings were collected from Parkinson's disease (PD) patients and healthy individuals in a controlled hospital environment to ensure high-quality data. All recordings were performed at the Neurology Department of Fırat University Hospital using the H1N Zoom microphone. To ensure consistency, the microphone was securely placed on a tripod and adjusted according to each participant's sitting position. A standard distance of 10 cm was maintained between the microphone and the participants during all recordings, as determined by reviewing the relevant literature.

We performed the extraction and visualization of various features of the recorded speech signals in MATLAB. Comparative evaluations were performed between three groups: PD patients and healthy individuals in MED ON and MED OFF conditions. A number of features, including jitter, shimmer, spectral entropy, skewness, kurtosis, and power bandwidth, were analyzed to identify patterns and differences between the groups.

MATLAB's Classification Learning Toolbox was used to evaluate the classification performance, allowing multiple machine learning algorithms to be tested. The extracted features were collectively introduced to the algorithms to evaluate their combined effectiveness in distinguishing groups. 15% of the dataset was reserved for testing.

### 2.1. Dataset

The study included voice recordings from 28 healthy individuals, 22 Med off patients, and 30 Med on patients. This distribution ensures a balanced dataset, allowing for robust comparisons between the groups and a comprehensive evaluation of the effects of pharmacological treatment on vocal characteristics in Parkinson's disease.

Participants were instructed to read aloud a predefined text, "Jale'nin Dünyası," to create a consistent dataset. The dataset was divided into three groups: PD patients in MED ON and MED OFF conditions, and healthy controls. To ensure uniformity in data collection, the same recording procedure was applied to all participants, including healthy individuals.

The speech samples were recorded and stored in WAV format and then transferred to a computer for analysis using MATLAB. The WAV format was chosen for its uncompressed and lossless characteristics, which preserve data integrity during storage and processing. This approach ensured that the recordings were suitable for detailed speech analysis and classification tasks.

**Table 1.** Dataset

| No | Class | Number of participants | |
|----|-------|------------------------|---|
| 1 | Without medication (Med Off) | 22 | |
| 2 | Medicated (Med On) | 30 | |
| 3 | Healthy Control | 28 | Total:80 |

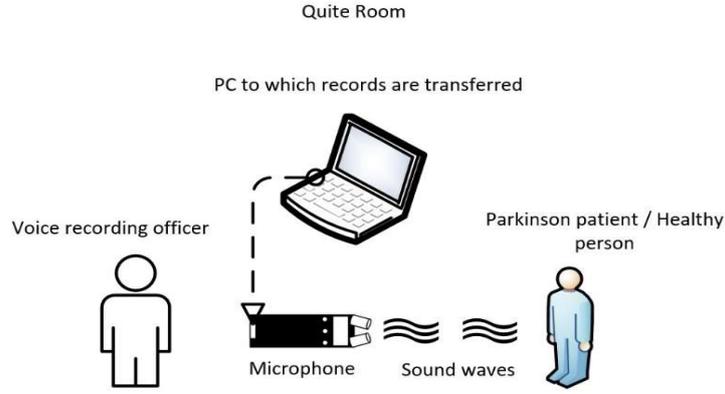

**Figure 1:** Experimental Setup

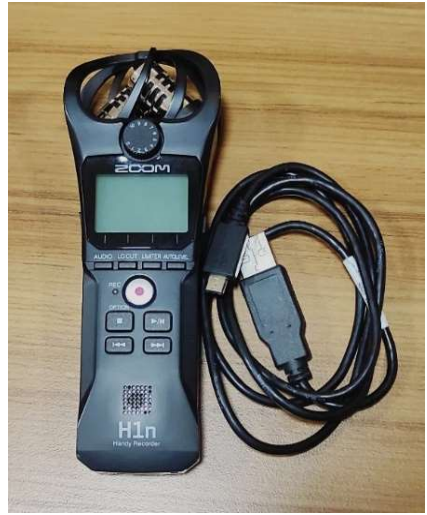

**Figure 2:** H1N Zoom microphone

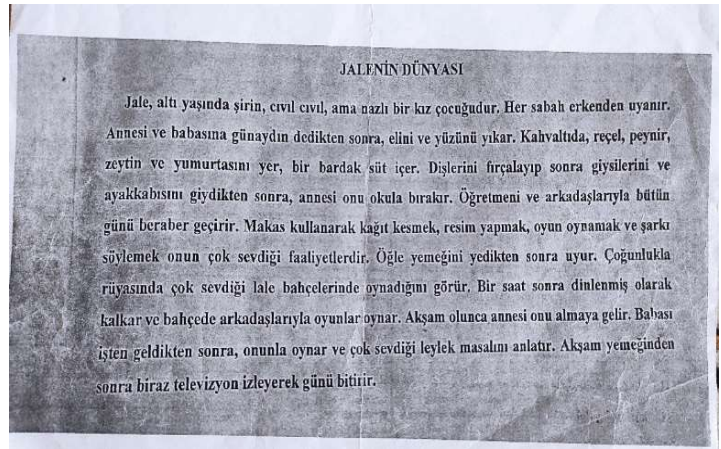

**Figure 3.** The text "Jale'nin Dünyası" read to healthy controls and Parkinson's Disease

## 3. FEATURES OBTAİNED FROM SPEECH SİGNALS

To effectively compare speech signals, it is essential to extract certain features from the speech. However, not all features are equally effective in distinguishing between signals. Therefore, it is crucial to identify and extract features that enhance the differentiation process and provide a more accurate and meaningful comparison.

**Table 2.** The used statistical moments.

| No | Feature | No | Feature | No | Feature | No | Feature |
|---|---|---|---|---|---|---|---|
| 1 | Maximum | 6 | Power Band Width | 11 | Variance | 16 | Entropy |
| 2 | Mean Frequency | 7 | Jitter | 12 | Amplitude Mean | 17 | ZCR |
| 3 | Minimum | 8 | Mean Energy | 13 | Median | 18 | Sure Entropy |
| 4 | Shimmer | 9 | Root mean square | 14 | Skewness | 19 | Q3-Q1(Interquartile Range) |
| 5 | Log Entropy | 10 | Standard deviation | 15 | Kurtosis | | |

### 3.1. Jitter – Shimmer

Jitter and shimmer are key parameters used to characterize speech signals and analyze distortions in speech. Jitter refers to variations in the fundamental frequency across consecutive cycles, reflecting frequency instability in the signal. Such variations are commonly associated with speech distortion and can contribute to perceptual qualities such as muffled, harsh, or rough sound. These parameters are typically obtained by analyzing the periodicity and determining the pitch through the autocorrelation method, which identifies each cycle of vocal cord vibration (Upadhya, Cheeran and Nirmal, 2017).

$$\frac{1}{N-1}\sum_{i=1}^{N-1}|P_i - P_{i+1}| \qquad (1)$$

The equation shown in (1) gives the formula that calculates the jitter value as a percentage. Jitter and shimmer are typically observed and analyzed together, as they are closely related parameters. One key point to note is that irregularities in the periodicity of signals are often linked to changes in energy, which reflect underlying inconsistencies in signal production. The equation shown in (2) gives the formula to calculate the shimmer value as a percentage.

$$\frac{1}{N-1}\sum_{i=1}^{N-1}\left|20\log\left(\frac{V_{i+1}}{V_i}\right)\right| \qquad (2)$$

### 3.2. Average Frequency

The mean frequency is an average frequency calculated as the sum of the product of the signal power spectrum and the frequency divided by the sum of the spectrum density(Altın and Er, 2016).

### 3.3. Average Energy

Average Energy - time graphs show the average energy distribution of speech signals over time. This analysis allows us to gain insight into the content of audio data from Parkinson's disease and healthy individuals. High energy regions may represent speech or specific sound events, while low energy regions may reflect silence or background noise.

### 3.4. Kurtosis

Kurtosis is a statistical metric that evaluates the extent of peakedness or flatness of a distribution in relation to its shape. It is commonly compared to the kurtosis of a normal distribution, which serves as a reference point for symmetric distributions. Higher kurtosis indicates sharper peaks, while lower kurtosis suggests a flatter distribution.

### 3.5. Skewness

Skewness is a statistical measure that quantifies the asymmetry of a distribution with respect to its axis of symmetry. It provides a numerical assessment of the degree and direction of deviation from the center point of the distribution.

### 3.6 Root Mean Square

Root Mean Square (RMS) is a fundamental metric used to quantify the power or amplitude of an speech signal. It provides a statistical measure of the signal's magnitude by calculating the square root of the mean of the squared values of the signal over a defined time period. RMS is particularly valuable in speech analysis as it represents the effective energy of the signal, offering insights into its overall intensity and variability.

$$\sqrt{\frac{1}{B}\sum_{i=1}^{b} x_i^2} \qquad (3)$$

It is found with the equation given in (3). B represents the total number of samples, x_i represents the signal, and i represents the sample

### 3.7. Zero Crossing Rate

Separation of the speech signal from moments of silence or other speech signals is very important in voice or speaker recognition systems. Zero crossing ratio is the most well-known method for sound level detection.(Ekim et al., 2008)

$$Z_S(m) = \frac{1}{L} \sum_{n=m-L+1}^{m} \frac{|sgn(s_1(n)) - sgn(s_1(n-1))|}{2} \qquad (4)$$

$$sgn(s_1(n)) = \begin{Bmatrix} +1, \ldots \ldots \ldots s_1(n) \geq 0 \\ -1, \ldots \ldots \ldots s_1(n) \leq 0 \end{Bmatrix} \qquad (5)$$

The zero crossing rate can be formulated as follows. Here L is the number of samples in each audio section. In a two-dimensional coordinate system, speech signals repeatedly intersect the x-axis; the y-values fluctuate in positive and negative ranges.

### 3.8. Entropy

Entropy is an analysis method in speech analysis that reveals the irregularity of a sound. Entropy has been applied in voice activity detection to identify silent and vocal speech regions. The discriminatory nature of this feature leads to its use in speech recognition. Entropy can be used to capture shapers or peaks of a distribution(Toh, Togneri and Nordholm, 2005).

### 3.9 Power Bandwidth

Power bandwidth analysis is conducted to evaluate the distribution of power across the frequency components of an speech signal. This process involves calculating power levels within specific frequency bands to identify the signal's spectral content. Speech signals are commonly analyzed by converting them from the time domain to the frequency domain through methods like the Fast Fourier Transform (FFT) or comparable algorithms. This transformation facilitates a detailed examination of the signal's frequency characteristics and power distribution.

### 3.10 Maxımum

Maximum value analysis provides an important quantitative indicator for seeing the change of sound signals over time. It expresses the highest amplitude in each segment.

## 4. RESULTS AND DİSCUSSİON

In this section, graphical representations of various features extracted from recorded speech signals using MATLAB will be presented. These plots provide a visual understanding of the differences in acoustic features between three groups: PD patients and healthy controls in MED ON and MED OFF conditions. Basic features such as jitter, shimmer, spectral entropy, root mean square energy, skewness, kurtosis, zero crossing ratio and power bandwidth were analyzed. The visualizations aim to highlight the variations and trends observed in the data and provide insights into how these features can distinguish PD patients from healthy individuals. For example, jitter and shimmer plots show irregularities in frequency and amplitude, while spectral entropy reflects the complexity of the speech signal. These graphical analyses form the basis of the subsequent classification phase. Following the feature analysis, the classification results obtained using MATLAB's Classification Learning Toolbox will be discussed. Decision Trees, Support Vector Machines, k-Nearest Neighbor, Naive Bayes, and Neural Network machine learning algorithms were applied to assess the diagnostic potential of the extracted features. The accuracy rates, sensitivity, and specificity of these models will be presented to evaluate their effectiveness in distinguishing Parkinson's patients from healthy individuals.

Show in Figures 4-5-6, the analysis of voice signals from healthy individuals demonstrates significantly fewer irregularities compared to those observed in the MED OFF and MED ON conditions of Parkinson's patients. The findings suggest that the tremors associated with Parkinson's disease contribute to pronounced irregularities in voice signals during speech. This irregularity is particularly evident in the parameters of voice tremor (jitter) and voice fluctuation (shimmer), which serve as key indicators of vocal instability. The comparative analysis highlights how these parameters differ across the groups, reflecting the impact of Parkinson's disease on vocal quality.

Moreover, this study underscores the potential of jitter and shimmer analysis in assessing the efficacy of pharmacological treatments. Observing changes in these parameters under MED ON conditions provides valuable insights into the extent to which drug therapy mitigates voice irregularities, thereby improving overall voice quality. These findings highlight the potential of voice signal analysis as a non-invasive method for monitoring disease progression and evaluating treatment outcomes in Parkinson's patients.

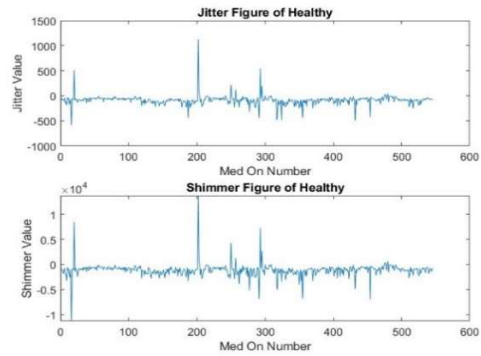

**Figure 4.** Jitter - shimmer figures of healthy controls

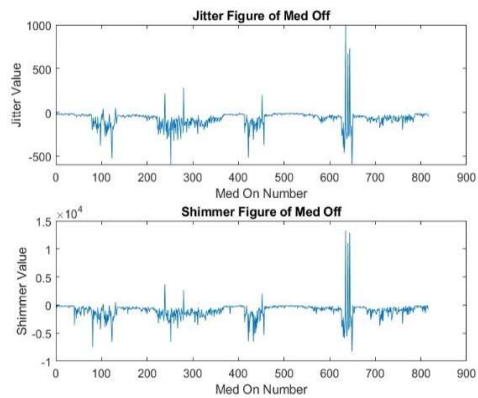

**Figure 5.** Jitter - shimmer figures of Med Off

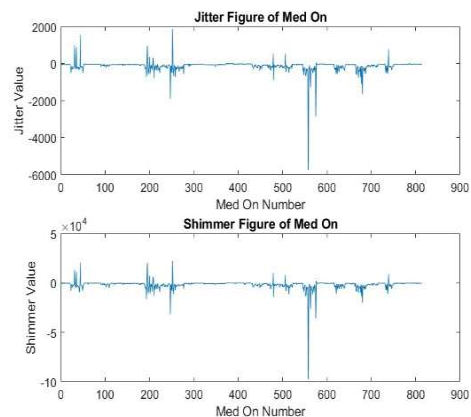

**Figure 6.** Jitter - shimmer figures of Med On

Show ın Figures 7-8-9, the amplitudes of healthy, med off and med on signals on the time axis are shown. It is seen that the amplitude values of the healthy signal have a larger amplitude value than the signal of the patient individual. It is understood that the patient signal cuts the 0 axis more than the healthy signal and has a more fractured structure, while the healthy signal draws a smoother graph. In addition, the effects of the effectiveness of pharmacological treatment methods used to alleviate tremor symptoms caused by Parkinson's disease on the sound parameters are seen

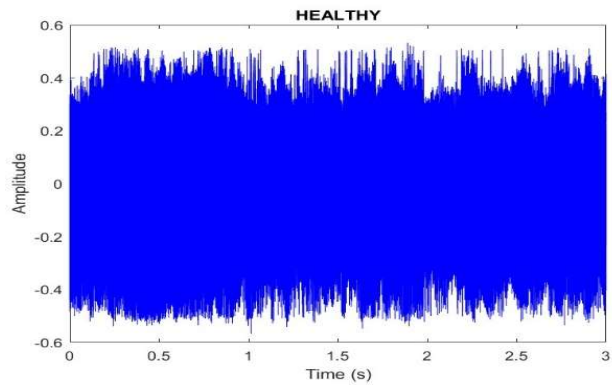

**Figure 7.** Amplitude of healthy signals on the 3s time

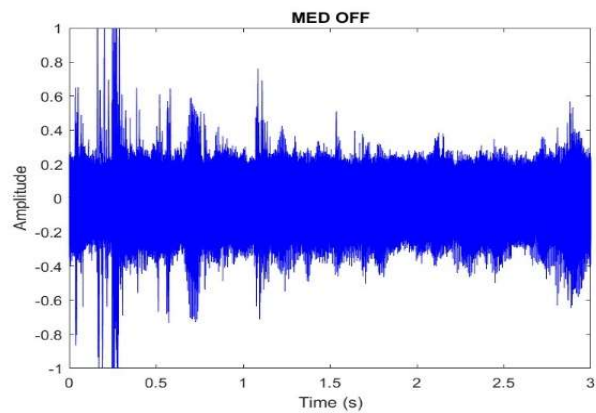

**Figure 8.** Amplitude of Med Off signals on the 3s time

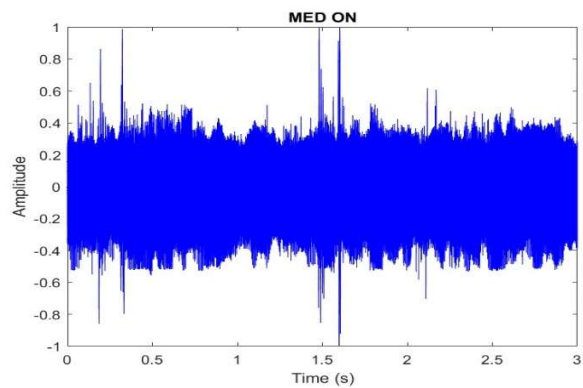

**Figure 9.** Amplitude of Med On signals on the 3s time

Show in Figures 10-11-12 the time-varying average frequency values for healthy individuals, Parkinson's patients in the Med off state, and Parkinson's patients in the Med on state. The voice frequencies of healthy individuals typically range between 500-600 Hz, exhibiting a more stable and regular pattern. This stability reflects better vocal control in healthy individuals. In contrast, the frequencies of Parkinson's patients in the Med off state are lower and more irregular compared to healthy individuals. These lower and less consistent frequency values highlight the adverse effects of Parkinson's symptoms on voice production. In the Med off state, patients' voice frequencies generally fall within the 300-400 Hz range, indicating the impact of the disease on motor control. For Parkinson's patients in the Med on state, the frequencies remain lower than those of healthy individuals but show noticeable improvement compared to the Med off state. This improvement demonstrates the positive effects of pharmacological treatment on vocal control. In summary, healthy individuals exhibit better vocal control than Parkinson's patients in the Med on state, and Parkinson's patients in the Med on state demonstrate better vocal control than those in the Med off state. These findings clearly highlight the beneficial effects of medication on voice frequency regulation.

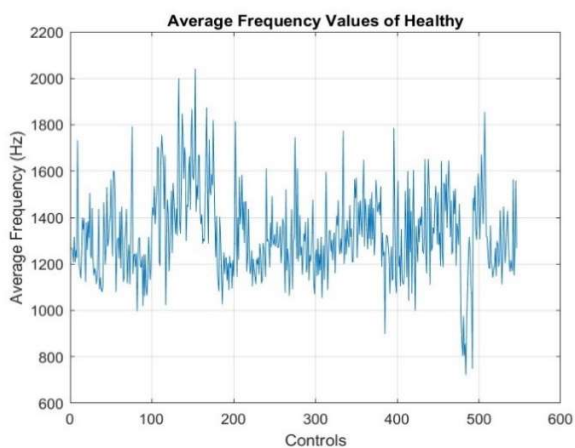

**Figure 10.** Average Freq. Values of Healthy

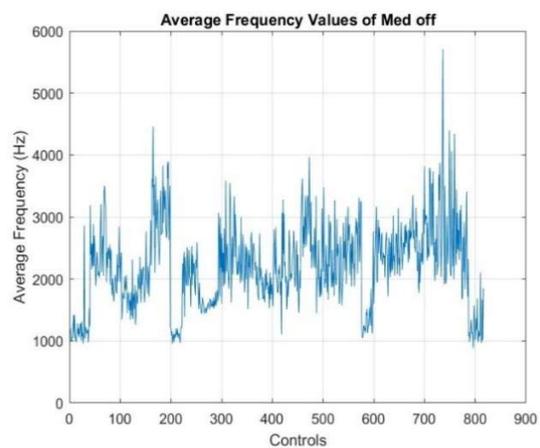

**Figure 11.** Average Freq. Values of Med off

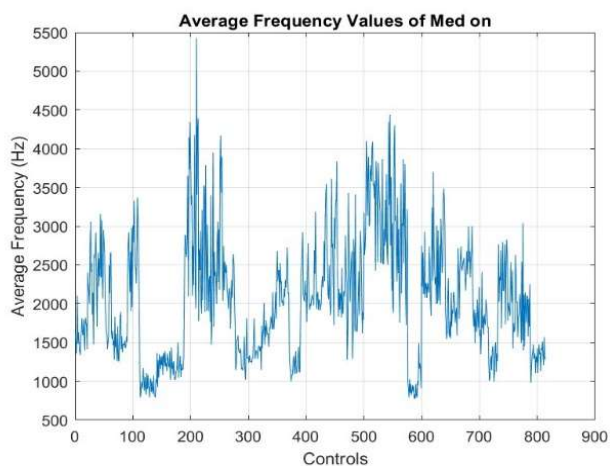

**Figure 12.** Average Freq. Values of Med on

Figures 13-14-15 compare the kurtosis characteristics of voice signals between individuals with Parkinson's and those without, highlighting the effects of pharmacological treatment. The results reveal a significant difference between healthy individuals and those in the Med off state, with kurtosis

values in the Med off condition being notably higher than those of healthy individuals. Furthermore, when comparing the Med on and Med off states, it is observed that kurtosis values are lower in the Med on state. This indicates a reduction in the irregularities and extreme variations in the voice signals following medication, reflecting the positive impact of pharmacological treatment.

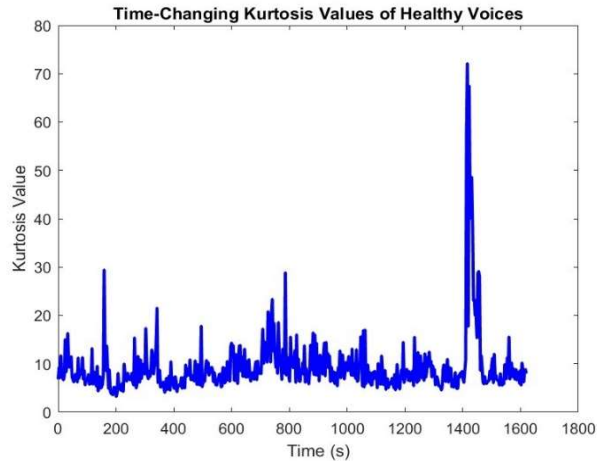

**Figure 13.** Kurtosis Values of Healthy Signals

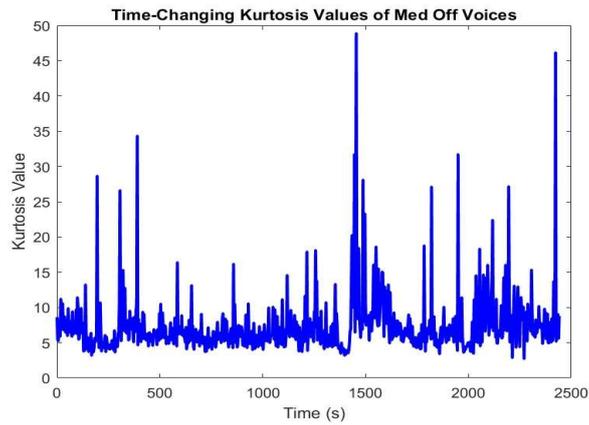

**Figure 14.** Kurtosis Values of Med Off Signals

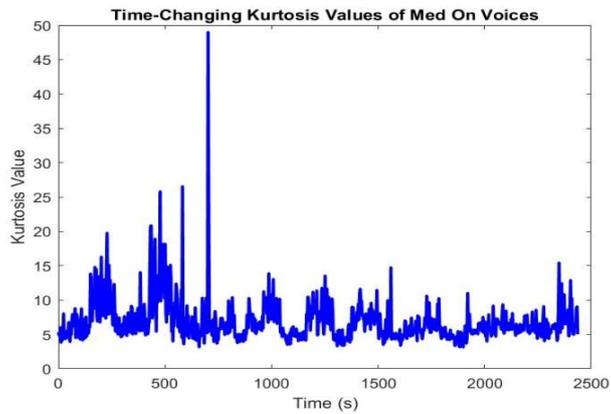

**Figure 15.** Kurtosis Values of Med On Signals

Skewness quantifies the asymmetry of a distribution relative to its axis of symmetry, providing a numerical evaluation of the deviation from symmetry around the distribution's center. Show in Figures 16-17-18, the skewness properties of the tree signals exhibit distinct variations under each condition.

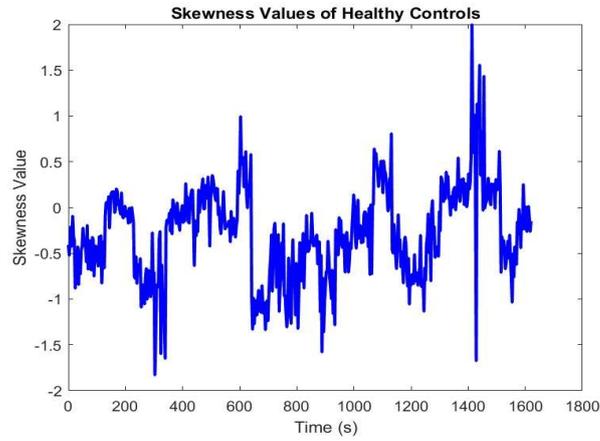

**Figure 16.** Skewness Values of Healthy Controls

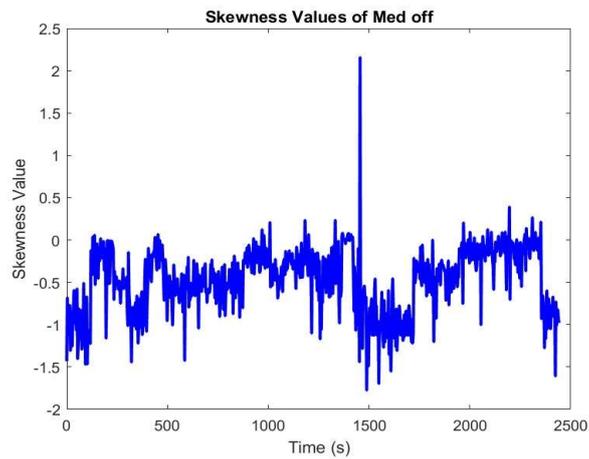

**Figure 17.** Skewness Values of Med off

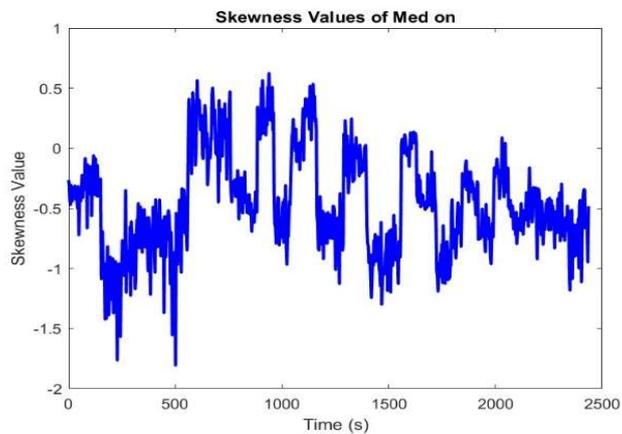

**Figure 18.** Skewness Values of Med On

# 5. CONCLUSİON

This study aimed to detect PD using speech signals. A separate speech dataset was collected covering three classes: medicated PD, without medicated PD and healthy control. 19 features were extracted from the speech signals. CHi2 feature selection algorithm was used due to its high score. 10-fold cross-validation technique was used to evaluate the model.

The extracted features were trained holistically and tested across various algorithms. Classification algorithms such as Knn, Support Vector Machine, Decision Tree, Naïve Bayes and Neural Network algorithms were used.

**Table 3.** Results of algorithms

| Classification algorithm | MED OFF-0 | HEALTHY-1 | MED ON-2 |
|---|---|---|---|
| **Decision Tree** | 74.8 | 80.5 | 61.3 |
| **Naive Bayes** | 78.3 | 82.1 | 40.4 |
| **SVM** | **82.4** | 85.4 | 75.7 |
| **k-NN** | 81.7 | **86.2** | 70.9 |
| **Neural Network** | 81.2 | 85.2 | **77.3** |

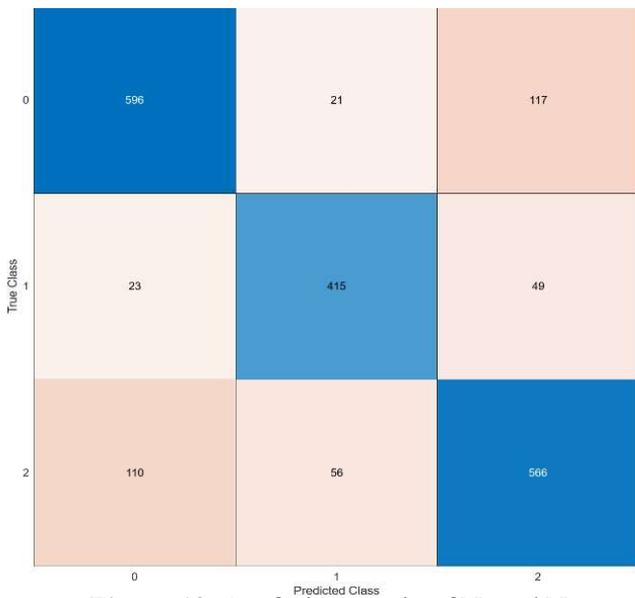

**Figure 19.** Confusion matrix of Neural Network

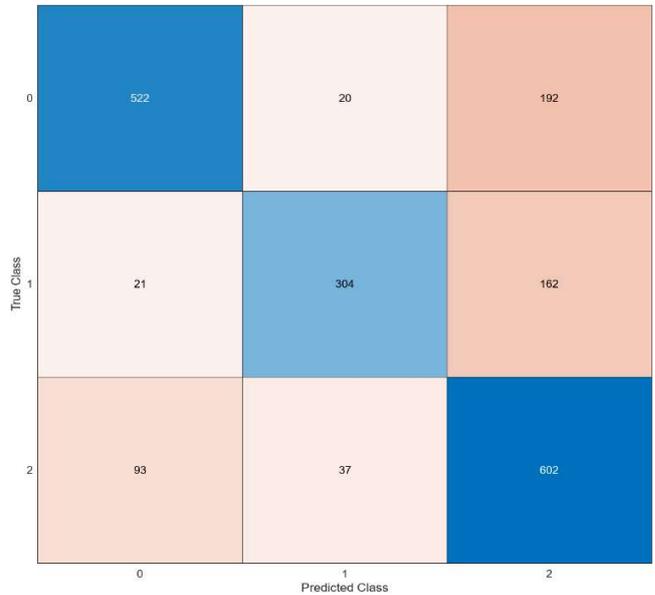

**Figure 20.** Confusion matrix of SVM model

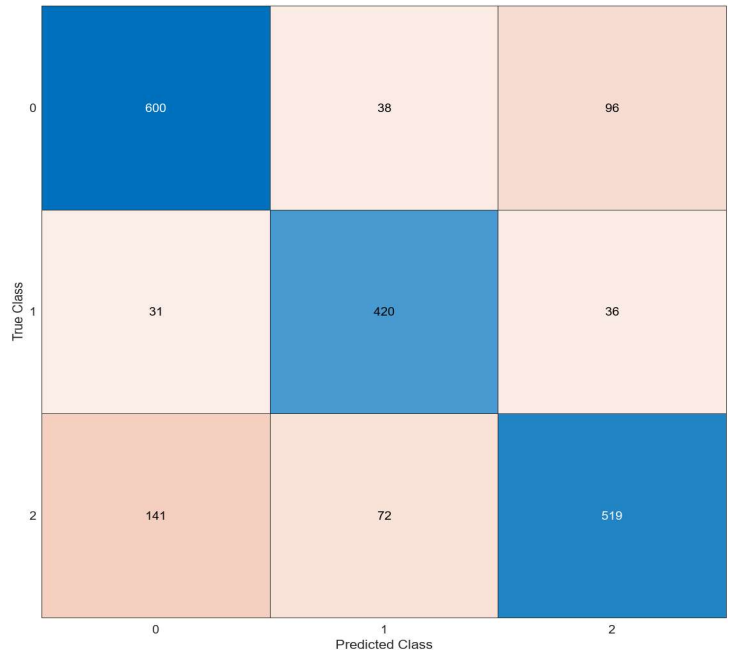

**Figure 21.** Confusion matrix of k-NN model

Show in figures 19-20-21, the confusion matrix of the 3 classification algorithms that make high predictions is presented.

**Table 4.** Confusion matrix parameters of 3 cases for Neural Network

| Class Name | Precision (%) | Recall (%) | F1-score (%) |
|---|---|---|---|
| 0 (Med Off) | 0.8120 | 0.8176 | 0.8148 |
| 1 (Healthy) | 0.8522 | 0.8435 | 0.8478 |
| 2 (Med On) | 0.7827 | 0.7827 | 0.7827 |
| Accuracy | 0.8106 | | |

**Table 5.** Confusion matrix parameters of 3 cases for SVM

| Class Name | Precision (%) | Recall (%) | F1-score (%) |
|---|---|---|---|
| 0 (Med Off) | 0.7112 | 0.8202 | 0.7620 |
| 1 (Healthy) | 0.6242 | 0.8421 | 0.7170 |
| 2 (Med On) | 0.8224 | 0.6297 | 0.7133 |
| Accuracy | 0.7312 | | |

**Table 6.** Confusion matrix parameters of 3 cases for k-NN

| Class Name | Precision (%) | Recall (%) | F1-score (%) |
|---|---|---|---|
| 0 (Med Off) | 0.8174 | 0.7772 | 0.7968 |
| 1 (Healthy) | 0.8624 | 0.7925 | 0.8260 |
| 2 (Med On) | 0.7090 | 0.7972 | 0.7505 |
| Accuracy | 0.7880 | | |

SVM showed high performance in identifying Med off. For healthy controls, all algorithms performed relatively well, with kNN achieving the best results. In Med on, the neural network outperformed the other algorithms. The relatively low accuracy rate performance in Med on can be attributed to the variability introduced by pharmacological treatments that significantly change patients' speech characteristics.

According to the confusion matrix parameters we obtained for the neural network, SVM and kNN, our neural network model achieved the highest accuracy for all cases.

## ACKNOWLEDGMENTS

This study has been produced from the master thesis of Burak ÇELİK.

## CONTRİBUTİONS OF THE AUTHORS

The authors' contributions to the paper are equal.

## CONFLİCT OF INTEREST STATEMENT

There is no conflict of interest between the authors.